\pdfoutput=1
\documentclass{article} %
\let\lstmemArticleAddContentsLine\addcontentsline
\usepackage{iclr2027_conference,times}
\let\addcontentsline\lstmemArticleAddContentsLine
\usepackage[hidelinks]{hyperref}
\usepackage{booktabs}
\usepackage{multirow}
\usepackage{graphicx}
\usepackage{wrapfig}
\usepackage{amsmath,amssymb}
\usepackage[table]{xcolor}
\usepackage{enumitem}
\iclrfinalcopy %
\newcommand{\lstmem}{LSTMem}
\newcommand{\dmem}{$\delta$-Mem}

\title{\raggedright\lstmem{}: Hierarchical Long Short-Term Online Memory for Large Language Models}

\author{%
  Xianglong Shi$^{1}$\thanks{Equal contribution.\quad$^{\dagger}$Corresponding authors.}\hphantom{$^{*}$}, Ruijie Yang$^{1*}$, Sirui Zhao$^{1\dagger}$, Shukang Yin$^{1}$, \\
  \textbf{Zihao Bian$^{1}$, Tinghao Yi$^{1}$, Enhong Chen$^{1\dagger}$} \\
  $^{1}$University of Science and Technology of China \\
  \texttt{xlshi@mail.ustc.edu.cn}
}

\begin{document}
\maketitle
\ificlrfinal\lhead{\lstmem{}: Hierarchical Long Short-Term Online Memory for Large Language Models}\fi
\addtocontents{toc}{\protect\setcounter{tocdepth}{-1}}

\begin{abstract}
Large language models increasingly serve as long-horizon assistants and agents, where they must both accumulate information across interactions and make the relevant parts available when later requests depend on them.
Existing compact online memories typically use a single persistent state both to accumulate history and to serve readout, so what the memory stores cannot be controlled separately from what it exposes to the current computation.
We propose LSTMem, an LSTM-inspired online memory that instead equips each layer of a frozen LLM with two matrix-valued states: a cell state that accumulates history and a hidden state whose readouts correct the backbone's attention. Input and forget gates control what the cell stores, while an output gate separately controls what the cell exposes through the hidden state.
LSTMem further connects memory across depth through forward hidden-state propagation and block-end feedback, and uses higher-layer reconstruction gradients to refine shallower-layer cell states before rebuilding hidden states from shallow to deep layers.
Across memory benchmarks on Qwen3-4B-Instruct, LSTMem consistently improves MemoryAgentBench, LoCoMo, and HotpotQA over the plain backbone. Comparisons further show that the LSTM-based memory formulation outperforms an associative-memory counterpart, while removing cross-layer hidden-memory propagation degrades performance.
These results demonstrate the benefits of separating memory accumulation from memory expression and organizing memory hierarchically across model depth.
The code is available at \href{https://github.com/Longchentong/LSTMem}{\textcolor{magenta}{this url}}.
\end{abstract}

\vspace{-4mm}
\section{Introduction}
\vspace{-1mm}
\label{sec:intro}

Long-horizon agents \citep{park2023generative,yang2024sweagent} and assistants \citep{packer2023memgpt,zhong2024memorybank} require language models to retain and reuse information across interactions \citep{maharana2024locomo,wu2025longmemeval,hu2025memoryagentbench,zhang2025agentsurvey}.
A natural approach is to include more history in the input, yet longer contexts can reduce performance \citep{liu2024lost,hsieh2024ruler,laban2025lost} even when the relevant evidence is correctly retrieved \citep{du-etal-2025-context}.
This motivates compact online memory \citep{munkhdalai2024infini,wang2024memoryllm,behrouz2025titans,lei2026deltamem}, a fixed-size state that incorporates new information during inference and persists across interactions \citep{zhoubian2026memorylargelanguagemodels}.
Such a memory must accumulate useful historical information and expose a suitable representation of it to the current computation.

\begin{wrapfigure}{R}{0.49\textwidth}
    \centering
    \vspace{-4mm}
    \includegraphics[width=\linewidth]{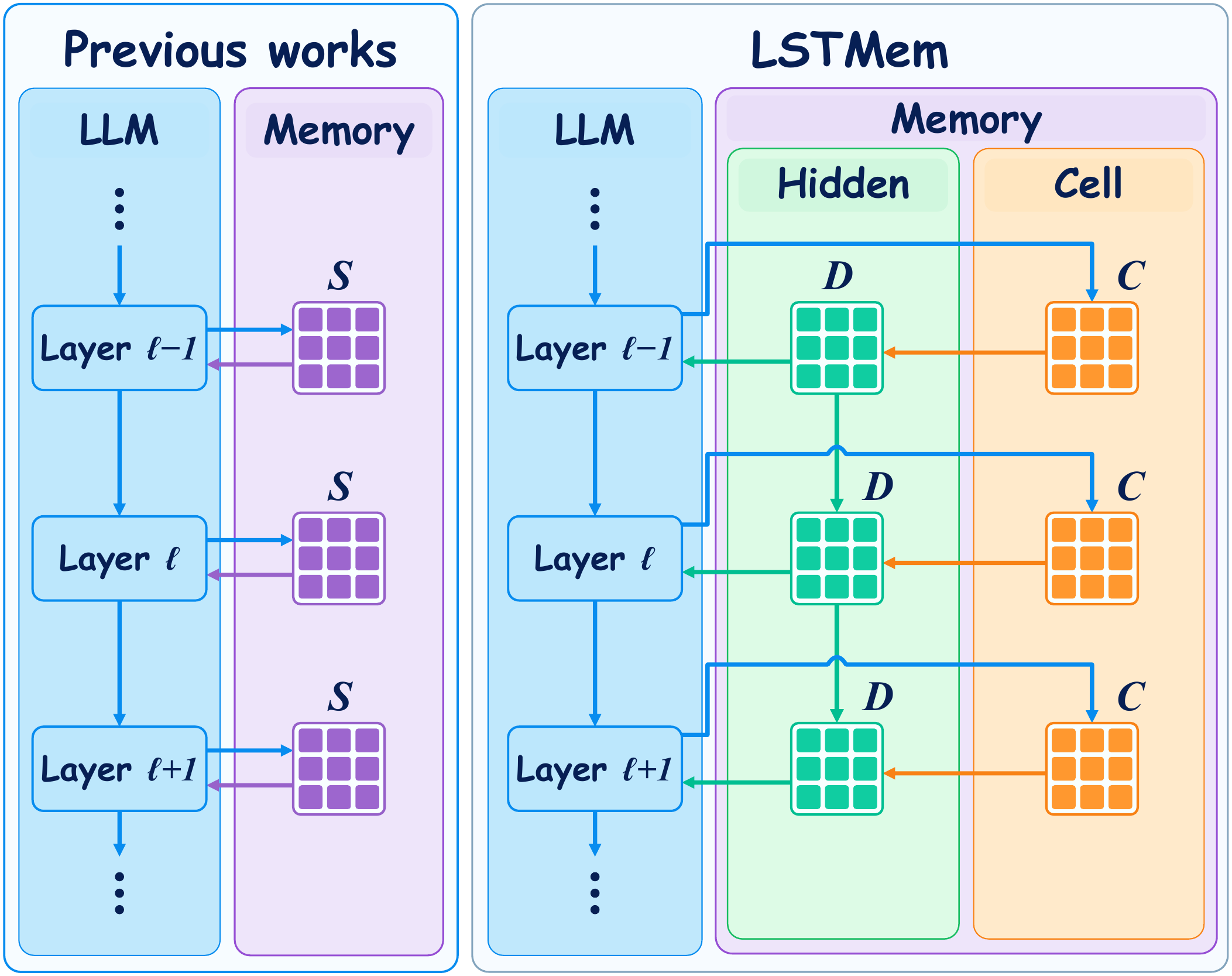}
    \vspace{-7mm}
    \caption{\textbf{Comparison of memory designs.}
    LSTMem separates accumulation in the cell state $C$ from memory expression in the hidden state $D$, and passes $D_t^{(\ell-1)}$ to layer $\ell$.
    }
    \label{fig:memory_overview}
    \vspace{-5mm}
\end{wrapfigure}

Existing compact online memories \citep{behrouz2025titans,lei2026deltamem} often build on the recurrent matrix states of linear attention \citep{peng2024eagle,dao2024transformers,qin2024hgrn2,yang2024deltanet,yang2025gated,kimiteam2025kimilinear}. Such states absorb long histories at a constant memory cost, but the same matrix both accumulates history and is exposed to the current computation.
For example as shown in Figure~\ref{fig:memory_overview}, recent $\delta$-Mem maintains an associative state $S_t$ within each memory head, updates it through a gated delta rule \citep{yang2025gated}, and reads from $S_{t-1}$ to produce history-dependent corrections to attention \citep{lei2026deltamem}. This is an effective way to couple online memory with a pretrained backbone, but it leaves accumulation and expression tied to the same state.
More generally, a representation well suited for persistent accumulation need not be the one that should be exposed to the current computation, because content kept for later can still affect the readout while it is irrelevant to the current step \citep{schlag2021linear,yang2024deltanet}.
Long short-term memory provides a natural organizing principle for this distinction, as an LSTM separates a cell state that accumulates information from a hidden state that exposes a controlled transformation of that accumulated content \citep{hochreiter1997lstm}. We take this separation between memory accumulation and memory expression as the starting point for our design.

Motivated by this principle, we propose LSTMem, a long short-term online memory for pretrained language models (Figure~\ref{fig:memory_overview}). First, LSTMem separates memory accumulation from memory expression, rather than tying both roles to the same recurrent state. Each memory module maintains a cell state $C$ that persistently accumulates historical information and a hidden state $D$ that exposes a gated representation of the accumulated memory to the current computation, with separate input, forget, and output gates controlling writing, retention, and expression. During our preliminary investigation, we further observed through layer-wise knowledge editing that manipulating memory-related information at different depths of an LLM can lead to substantially different downstream effects in Appendix~\ref{app:layer_editing}. Therefore, we argue that memory in a deep language model should not be treated merely as a collection of independent, layer-local states and introduce a layer-hierarchical memory mechanism in which the hidden memory $D_t^{(\ell-1)}$ from a shallower layer is propagated to and conditions the memory at the next layer, influencing its state construction and subsequent readout. This forward path is complemented by block-end feedback, which uses deeper-layer reconstruction gradients to refine shallower-layer cell states at history-block boundaries.

We evaluate \lstmem{} on MemoryAgentBench \citep{hu2025memoryagentbench} and HotpotQA \citep{yang2018hotpotqa} using Qwen3-4B-Instruct, Qwen3-8B \citep{yang2025qwen3}, and SmolLM3-3B \citep{bakouch2025smollm3}, and further test conversational memory and general capabilities on LoCoMo \citep{maharana2024locomo}, IFEval \citep{zhou2023ifeval}, and GPQA-Diamond \citep{rein2023gpqa} with Qwen3-4B-Instruct.
Relative to the published backbone results, \lstmem{} improves MemoryAgentBench and HotpotQA F1 on all three models (Table~\ref{tab:main}).
On Qwen3-4B-Instruct, it gains $15.54$ points on MemoryAgentBench, $12.81$ on LoCoMo, and $12.12$ on HotpotQA F1; its IFEval score is only $0.19$ points lower and its GPQA-Diamond score $3.03$ points higher (Tables~\ref{tab:main}--\ref{tab:general_capabilities}).
Removing hidden-memory propagation across layers lowers MemoryAgentBench by $1.92$ and $1.77$ points on two training seeds, with the largest drops on test-time learning ($6.21$ and $8.00$ points; Table~\ref{tab:c2}).
These gains come at a modest cost, as \lstmem{} retains on average $92.10\%$ of the decoding speed of \dmem{} and takes only $5.9\%$ to $6.5\%$ longer end-to-end with 2,048-token answers (Appendix~\ref{app:efficiency}).

Our contributions are summarized as follows:
\vspace{-1mm}
\begin{itemize}[leftmargin=1.2em,itemsep=3pt,topsep=3pt]
\item We propose \lstmem{}, an online memory for frozen pretrained language models that uses separately gated states $C,D$ to accumulate history and supply low-rank attention corrections through hidden-memory readouts.
\vspace{-1mm}
\item We introduce a layer-hierarchical memory structure in which $D_t^{(\ell-1)}$ connects successive memory layers, together with block-end feedback that uses higher-layer reconstruction gradients to revise lower-layer cells and then rebuilds the hidden states.
\vspace{-1mm}
\item Extensive experiments on MemoryAgentBench, LoCoMo, and HotpotQA across three LLMs demonstrate the effectiveness of \lstmem{}. Comparisons further indicate that organizing memory hierarchically across depth is beneficial.
\end{itemize}
\vspace{-2mm}

\section{Preliminaries}
\label{sec:prelim}
\vspace{-2mm}
\textbf{Associative memory.}
At token position $t$, each memory head of width $r$ maintains a fixed-size state $S_t\in\mathbb{R}^{r\times r}$ that stores associations between keys and values $k_t,v_t\in\mathbb{R}^{r}$ \citep{yang2024deltanet,lei2026deltamem}.
Holding $k_t,v_t$ fixed, a gradient descent step on $\frac{1}{2}\|Sk_t-v_t\|_2^2$ with respect to $S$ at $S_{t-1}$ gives the delta-rule update
\begin{equation*}
S_t=S_{t-1}+\eta_t\bigl(v_t-S_{t-1}k_t\bigr)k_t^\top,
\end{equation*}
where $\eta_t\ge0$ is the scalar write rate.
The residual directs writes toward associations that the current state predicts incorrectly.

\textbf{LSTM accumulation and expression.}
LSTM separates accumulated content in a cell state $c_t$ from its expression through a hidden state $s_t$ \citep{hochreiter1997lstm,gers2000forget,beck2024xlstm}:
\begin{equation}
c_t=f_t\odot c_{t-1}+i_t\odot\widetilde c_t,\qquad
s_t=o_t\odot\tanh(c_t).
\label{eq:lstm_background}
\end{equation}
Here $\widetilde c_t$ is the candidate content and $\odot$ denotes elementwise multiplication.
The input and forget gates $i_t,f_t$ regulate writing and retention, while the output gate $o_t$ controls memory expression.
At layer $\ell$, \lstmem{} adopts this separation with matrix states $C_t^{(\ell)}$ for accumulation and $D_t^{(\ell)}$ for readout and propagation across layers (Section~\ref{sec:method}).

\section{LSTMem}
\label{sec:method}

Figure~\ref{fig:arch} shows how LSTMem reads memory to guide attention (Section~\ref{sec:read}) and updates memory through gated accumulation and expression (Section~\ref{sec:cell}).
We then describe cross-layer feedback for refining memory at block boundaries (Section~\ref{sec:block_feedback}; Figure~\ref{fig:feedback}).
Section~\ref{sec:recipe} presents the training and inference procedure. Further details are provided in Appendix~\ref{app:memory_impl}.

\begin{figure}[t]
\centering
\includegraphics[width=\linewidth]{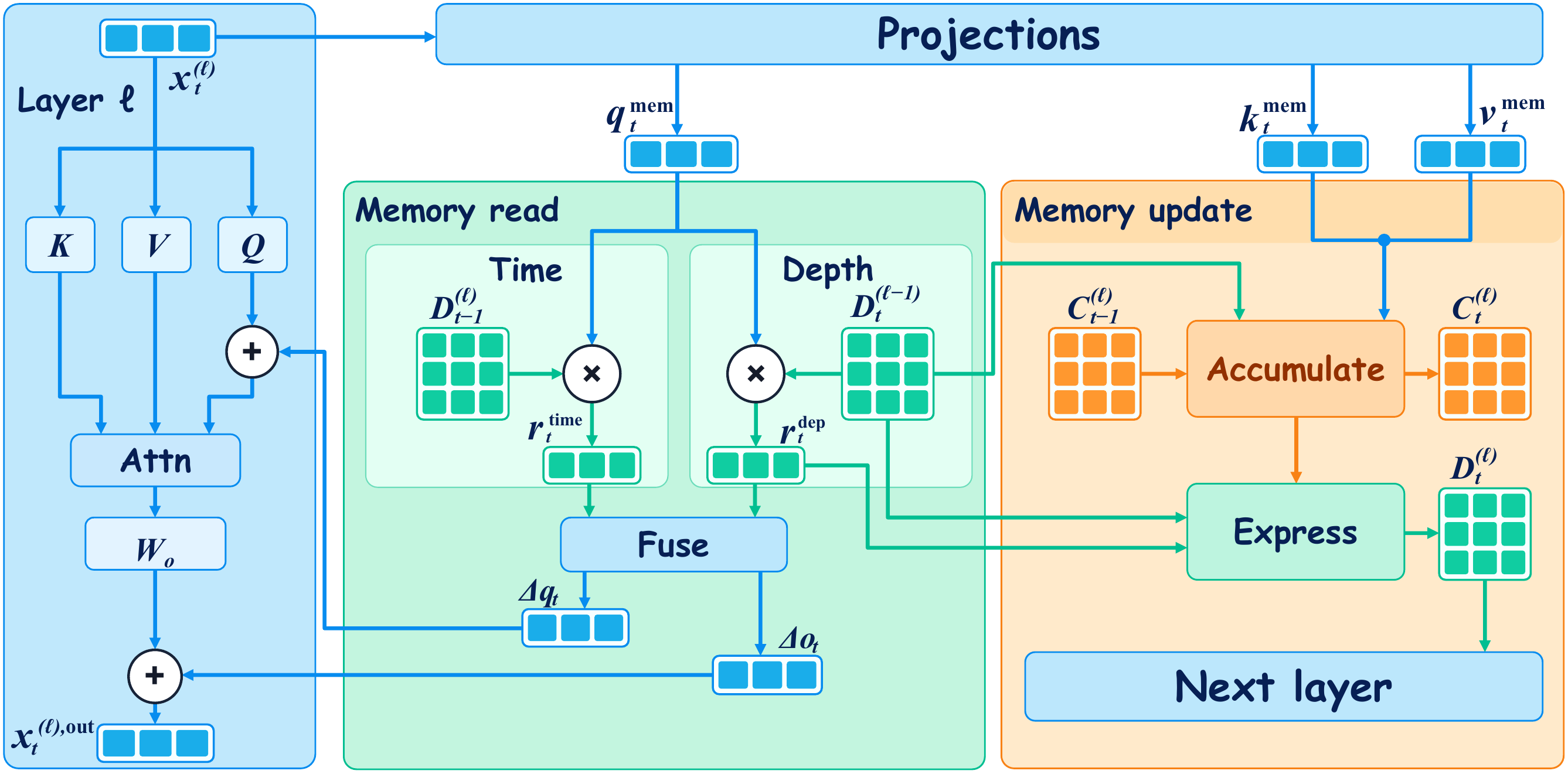}
\input{figures/editable/lstmem_read_update_caption.tex}
\end{figure}

\subsection{Memory readout and attention correction}
\label{sec:setting}
\label{sec:read}
\label{sec:attention}

Given the input $x_t^{(\ell)}\in\mathbb{R}^{d}$ to the attention module at position $t$ in layer $\ell$, LSTMem projects it to low-dimensional memory vectors $q_t^{\mathrm{mem}},k_t^{\mathrm{mem}},v_t^{\mathrm{mem}}\in\mathbb{R}^{d_s}$:
\begin{equation}
\begin{aligned}
q_t^{\mathrm{mem}}=\mathrm{n}\!\left(W_q^{\mathrm{mem}}x_t^{(\ell)}\right),\quad
k_t^{\mathrm{mem}}=\mathrm{n}\!\left(W_k^{\mathrm{mem}}x_t^{(\ell)}\right),\quad
v_t^{\mathrm{mem}}=W_v^{\mathrm{mem}}x_t^{(\ell)}.
\end{aligned}
\label{eq:projections}
\end{equation}

The vector $q_t^{\mathrm{mem}}$ reads memory, while $k_t^{\mathrm{mem}}$ and $v_t^{\mathrm{mem}}$ participate in the update in Section~\ref{sec:cell}.
Memory has $n_h$ heads of vector width $r$, so $d_s=n_hr$; $\mathrm{n}(\cdot)$ applies $\tanh$ followed by $\ell_2$ normalization within each head.
Memory projection parameters are specific to each layer. We omit layer superscripts on projections and intermediate vectors when describing a fixed memory layer.

Each layer maintains $C^{(\ell)},D^{(\ell)}\in\mathbb{R}^{n_h\times r\times r}$ for historical accumulation and memory expression.
Before writing the current position, the same $q_t^{\mathrm{mem}}$ reads two memories: $D_{t-1}^{(\ell)}$ carried from the preceding position in the current layer, and $D_t^{(\ell-1)}$ formed at the current position in the preceding layer.
These correspond to the temporal and depth read paths in Figure~\ref{fig:arch}.
Let $j\in\{1,\ldots,n_h\}$ index memory heads, with $[j]$ selecting the corresponding vector or matrix and $\operatorname{concat}_j$ concatenating head outputs:
\begin{equation}
\begin{gathered}
r_t^{\mathrm{time}}=\operatorname{concat}_{j}\!\left(D_{t-1}^{(\ell),[j]}q_t^{\mathrm{mem},[j]}\right),
\quad
r_t^{\mathrm{dep}}=\operatorname{concat}_{j}\!\left(D_t^{(\ell-1),[j]}q_t^{\mathrm{mem},[j]}\right),\\[2pt]
u_t=W_{\mathrm{read}}\bigl[r_t^{\mathrm{time}};r_t^{\mathrm{dep}}\bigr]+b_{\mathrm{read}}.
\end{gathered}
\label{eq:read}
\end{equation}

Fusing the two readout vectors lets the current layer use its own history together with memory newly formed at the same position in the preceding layer.
Using one-based layer indices, we set $D_t^{(0)}=\mathbf{0}$ as the depth input to the first memory layer.
Two projections map the fused signal $u_t\in\mathbb{R}^{d_s}$ to attention corrections:
\begin{equation}
\Delta q_t=\frac{\alpha}{r}W_{\Delta q}u_t,
\qquad
\Delta o_t=\frac{\alpha}{r}W_{\Delta o}u_t.
\label{eq:injection}
\end{equation}

The projection matrices $W_{\Delta q}$ and $W_{\Delta o}$ are learned, and $\alpha$ is a fixed scale constant.
We add $\Delta q_t$ to the model's original query before native query normalization and positional encoding \citep{su2024roformer}, allowing memory to influence access to the explicit context; keys and values follow the original computation.
After the output projection, we add $\Delta o_t$ to the attention result, letting memory directly correct the output representation.

\subsection{Gated accumulation and expression}
\label{sec:cell}
\label{sec:depth}

LSTMem uses $C$ to accumulate historical content and $D$ to provide the representation for reading and propagation across layers.
This separation allows content updates and memory expression to be controlled separately.

In the memory update illustrated in Figure~\ref{fig:arch}, lower-layer memory provides a reference for the current layer's write.
We read $D_t^{(\ell-1)}$ with $k_t^{\mathrm{mem}}$ to obtain $p_t^{\mathrm{dep}}$ and construct the write residual $e_t$ from its prediction:
\begin{equation}
p_t^{\mathrm{dep}}=\operatorname{concat}_{j}\!\left(D_t^{(\ell-1),[j]}k_t^{\mathrm{mem},[j]}\right),
\quad
\hat v_t=\widetilde W_{\mathrm{pred}}p_t^{\mathrm{dep}}+b_{\mathrm{pred}},
\quad
e_t=v_t^{\mathrm{mem}}-\hat v_t.
\label{eq:residual}
\end{equation}

The prediction readout $p_t^{\mathrm{dep}}$ comes from $k_t^{\mathrm{mem}}$.
The gates use $r_t^{\mathrm{dep}}$, read with $q_t^{\mathrm{mem}}$ in Section~\ref{sec:read}, together with the current input representation $z_t$ and write residual $e_t$.
The same layer input $x_t^{(\ell)}$ also produces $z_t\in\mathbb{R}^{d_s}$:
\begin{equation}
z_t=W_xx_t^{(\ell)}+b_x,
\quad
[i_t;f_t;o_t;g_t]=\sigma\!\left(\widetilde W_{\mathrm{gate}}[z_t;r_t^{\mathrm{dep}};e_t]+b_{\mathrm{gate}}\right).
\label{eq:gates}
\end{equation}

Here $\sigma$ is the elementwise sigmoid function, and $i_t,f_t,o_t,g_t\in(0,1)^{d_s}$ are the input, forget, output, and depth gates.
The input and forget gates control writing and retention in $C$; the output and depth gates control how local accumulation and lower-layer memory are expressed in $D$.
For each head, the two state updates are
\begin{align}
C_t^{(\ell),[j]}
={}&\operatorname{diag}\!\left(f_t^{[j]}\right)C_{t-1}^{(\ell),[j]}
+\operatorname{diag}\!\left(i_t^{[j]}\right)
e_t^{[j]}\left(k_t^{\mathrm{mem},[j]}\right)^\top,
\label{eq:write}\\[4pt]
D_t^{(\ell),[j]}
={}&\operatorname{diag}\!\left(o_t^{[j]}\right)
\tanh\!\left(C_t^{(\ell),[j]}\right)
+\operatorname{diag}\!\left(g_t^{[j]}\right)D_t^{(\ell-1),[j]}.
\label{eq:hidden}
\end{align}

Lower-layer memory thus participates both in constructing the write candidate and in forming the expressed state, establishing a continuous memory pathway across depth.
Updated $C$ and $D$ persist across positions, while $D$ also passes to the next layer at the same position.
Effective projection parameterizations and parallel computation within a block are detailed in Appendix~\ref{app:memory_impl}.

\subsection{Cross-layer feedback}
\label{sec:block_feedback}

We process history in consecutive segments, referred to as history blocks.
Within a block, memory flows from shallow to deep layers, so a lower layer cannot yet use higher-layer reconstruction errors for the same history when it writes.
LSTMem therefore introduces feedback at block boundaries to correct lower-layer cells using higher-layer errors (Figure~\ref{fig:feedback}).

\begin{figure}[!htbp]
\centering
\includegraphics[width=\linewidth]{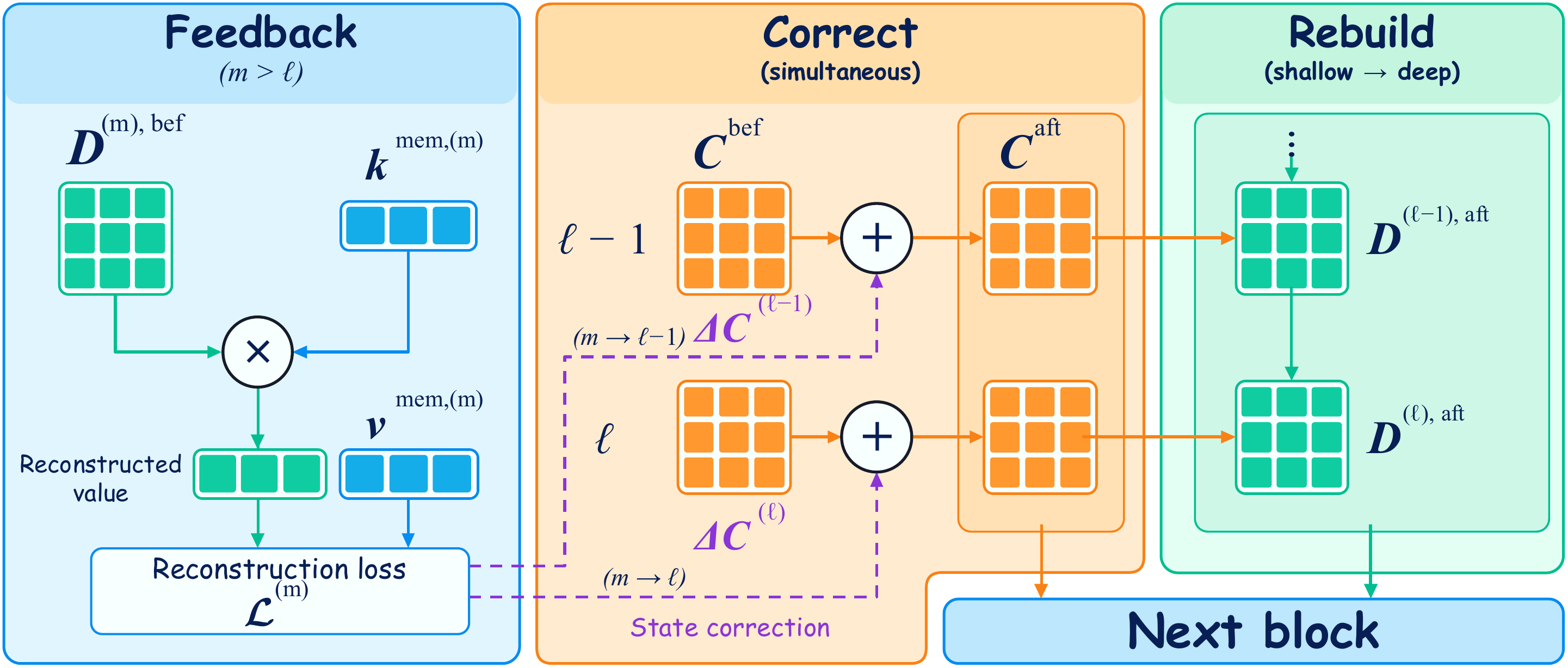}
\input{figures/editable/lstmem_block_feedback_caption.tex}
\end{figure}

Let $b$ index the current history block, and use superscripts $\mathrm{bef}$ and $\mathrm{aft}$ for terminal states before and after feedback.
All layers share a set of sampled positions $\mathcal I_b$ from this block.
Using the memory keys and values already computed at these positions, each layer reconstructs its associations from its terminal expressed state:
\begin{equation}
\mathcal L_b^{(\ell)}
=\frac{1}{2|\mathcal I_b|n_hr}
\sum_{t\in\mathcal I_b}\sum_{j=1}^{n_h}
\left\|
D_b^{(\ell),\mathrm{bef},[j]}
k_t^{\mathrm{mem},(\ell),[j]}
-v_t^{\mathrm{mem},(\ell),[j]}
\right\|_2^2.
\label{eq:feedback_loss}
\end{equation}

Associations from early positions are tested after later writes have modified the state, so the loss measures whether they remain recoverable at the block boundary.
Sharing $\mathcal I_b$ across layers ties all reconstruction errors to the same historical tokens, while each layer is tested on its own projected keys and values, so every loss reflects recall in that layer's own key--value space.

These historical associations provide feedback without additional answer labels.
For layer $\ell$, the correction uses reconstruction errors only from higher layers $m>\ell$.
The inner derivative treats terminal cell states as independent variables and follows only the depth-state recurrence in Eq.~\ref{eq:hidden}, with sampled keys, values, terminal gates, and the other terminal cell states held fixed.
With $L$ memory layers, the correction is summarized by
\begin{equation}
\begin{aligned}
G_b^{(\ell)}
=\frac{\partial}{\partial C_b^{(\ell),\mathrm{bef}}}
\sum_{m=\ell+1}^{L}\mathcal L_b^{(m)},\quad
C_b^{(\ell),\mathrm{aft}}
=C_b^{(\ell),\mathrm{bef}}+\Delta C_b^{(\ell)}.
\end{aligned}
\label{eq:feedback_step}
\end{equation}

The correction $\Delta C_b^{(\ell)}$ follows the negative direction of $G_b^{(\ell)}$, with headwise normalization, a fixed step size, and a cell RMS scale bounded below by one.
All corrections are evaluated at the same pre-feedback states and committed simultaneously after numerical checks.
The highest layer receives no error from above, so its cell remains unchanged.

Correcting $C$ also requires updating the expressed state $D$ used for reading and depth propagation.
Starting from the pre-feedback states and cached gates, we rebuild $D$ from shallow to deep layers along the existing depth connection.
Each rebuilt state incorporates both its local cell correction and the change in the preceding layer's expressed state.
The highest-layer $D$ can therefore change even though its cell is unchanged.
The corrected $C$ and $D$ initialize memory for the next history block.
Feedback changes only the memory carried forward; model parameters and outputs already computed within the block remain unchanged.
The gradient recurrence, headwise correction rule, and incremental rebuild are given in Appendix~\ref{app:feedback_impl}.

\subsection{Training and inference}
\label{sec:recipe}

LSTMem trains the adapter with answer cross-entropy while keeping the base model frozen.
Each training example starts from zero memory: we process its history, update $C$ and $D$, and apply block-end feedback before predicting the final assistant turn using the resulting states.
During the training answer pass, memory writing and feedback are disabled, and the historical KV cache is not carried into answer computation. Both read paths query the cached terminal hidden states of the corresponding memory layers, while $C$ and $D$ remain fixed across answer tokens.
The stored memory influences attention through the readout path in Section~\ref{sec:read}.
The reconstruction objective in Section~\ref{sec:block_feedback} corrects these states and is not added as an auxiliary term to the answer loss.

Training begins on QASPER \citep{dasigi2021qasper} with the history computation graph retained.
Answer-loss gradients can therefore pass through memory readout into historical writing and feedback computation, supervising how the memory is constructed as well as how it is used.
Holding keys, values, and gates fixed in Section~\ref{sec:block_feedback} applies only to the inner feedback derivative; it does not stop answer-loss gradients through these computations.
We then continue training on the Long dataset, which contains 71 synthetic histories ranging from 44,074 to 98,340 tokens (Appendix~\ref{app:training_call}). Histories in the Long dataset are still written and corrected in blocks, but history processing runs without gradient tracking, and states are detached between blocks.
For these examples, the answer loss trains the adapter through memory readout and attention correction.

At inference, all model parameters are fixed.
Historical writing and block-end feedback continue to form $C$ and $D$, whose readouts influence subsequent answers.
State updates during answer generation and the history retained in the prompt follow the corresponding evaluation protocols~\citep{lei2026deltamem}; memory is reset between independent conversations (Appendix~\ref{sec:eva}).
The training objective, training data, initialization, and evaluation settings are given in Appendices~\ref{app:protocol} and~\ref{app:training_call}.

\section{Experiments}
\label{sec:exp}

We evaluate \lstmem{} on MemoryAgentBench (MAB; \citealp{hu2025memoryagentbench}) and HotpotQA \citep{yang2018hotpotqa} using Qwen3-4B-Instruct, Qwen3-8B \citep{yang2025qwen3}, and SmolLM3-3B \citep{bakouch2025smollm3}.
On Qwen3-4B-Instruct, we also evaluate LoCoMo \citep{maharana2024locomo}, IFEval \citep{zhou2023ifeval}, and GPQA-Diamond \citep{rein2023gpqa}.
The main 4B model backbone is frozen, and only the memory adapter is trained.
The baseline results in Tables~\ref{tab:main}--\ref{tab:general_capabilities} are taken from \citet{lei2026deltamem}, with MAB averages recomputed from the published family scores using question-count weights.
LoCoMo averages are also weighted by question count. Table~\ref{tab:c1} additionally reports \dmem{} results after our second-stage training on the Long dataset.
We first present the main results on the memory benchmarks and on general capabilities in Section~\ref{sec:main} and then ablate the depth coupling, the choice of memory layers, the training stages, and block-end feedback in Section~\ref{sec:ablation}.
On Qwen3-4B-Instruct, \lstmem{} retains $92.10\%$ of \dmem{}'s equivalent single-step decoding speed on average. Its end-to-end times are $5.9\%$--$6.5\%$ longer than \dmem{}'s for 8K-, 32K-, and 128K-token histories with 2,048-token answers as detailed in Appendix~\ref{app:efficiency}.
More detailed training and evaluation settings are provided in Appendix~\ref{app:protocol}.

\textbf{Baselines.}
We compare \lstmem{} with four families of memory methods on the same backbones, namely external memory with BM25 RAG \citep{lewis2020rag,robertson2009bm25}, LLMLingua-2 \citep{pan2024llmlingua2,jiang2023llmlingua}, and MemoryBank \citep{zhong2024memorybank}; parametric memory with Context2LoRA \citep{hu2022lora,back2026context2lora} and MemGen \citep{zhang2025memgen}; auxiliary memory with MLP Memory \citep{wei2026mlpmemory}; and online memory with \dmem{} \citep{lei2026deltamem}.
For \dmem{}, we report its multi-state write (MSW) variant, which obtains the best MAB and LoCoMo averages among its variants on Qwen3-4B-Instruct and is also the most direct comparison, as it likewise keeps parallel $r\times r$ states per layer, corrects the attention query and output with the same $r$ and $\alpha$, and is trained on QASPER with the same settings as our first stage. Unlike \lstmem{}, MSW reads the same states it writes and has no depth coupling or block-end feedback.

\textbf{Implementation details.}
On Qwen3-4B-Instruct, \lstmem{} attaches memory to all 36 layers with $n_h=4$ heads of width $r=8$ and $\alpha=16$. The persistent $C/D$ state for one sequence contains 18,432 values, occupying 36 KiB in BF16 independently of history length.
Block-end feedback sums the reconstruction losses of all higher layers over at most 32 evenly spaced positions per history block and runs in both training and inference.
Both training stages use AdamW with batch size 32, BF16, and a cosine schedule with 10\% warm-up.
The first stage runs 70 updates on QASPER at a learning rate of $2\times10^{-4}$, with histories capped at 8,192 tokens.
The second stage runs 26 updates at $5\times10^{-5}$ with a fresh optimizer and writes each history in 2,048-token blocks, bringing the full schedule to only 96 updates. Further implementation and training details are given in Appendix~\ref{app:memory_impl}.

\subsection{Main results}
\label{sec:main}

\begin{table}[!t]
\centering
\caption{\textbf{Main results across three base models.}
Bold marks the highest score within each model group; subscripts give point changes from its unadapted model.
Baseline scores are from \citet{lei2026deltamem}.
HotpotQA is scored by exact match (EM) and token-level F1 against the gold answer.
AR, TTL, LRU and SF denote accurate retrieval, test-time learning, long-range understanding and selective forgetting.
AR and SF are scored by accuracy, TTL by classification accuracy and movie-recommendation Recall@5, and LRU by summarization F1 and Detective QA accuracy.
MAB averages use 2,000/700/171/800 question weights for AR/TTL/LRU/SF; baseline averages are approximated from rounded family scores.}
\label{tab:main}
\begingroup
\fontsize{8}{10}\selectfont
\setlength{\tabcolsep}{1.6pt}
\renewcommand{\arraystretch}{1.16}
\setlength{\heavyrulewidth}{.65pt}
\setlength{\lightrulewidth}{.35pt}
\resizebox{\linewidth}{!}{%
\begin{tabular}{@{}lccccccc@{}}
\toprule
\multirow{2}{*}{\textbf{Model}} & \multicolumn{2}{c}{\textbf{HotpotQA} $\uparrow$} & \multicolumn{5}{c}{\textbf{MemoryAgentBench} $\uparrow$} \\
\cmidrule(lr){2-3}\cmidrule(l){4-8}
 & EM & F1 & \textbf{Avg.} & AR & TTL & LRU & SF \\
\midrule
\textbf{Qwen3-4B-Instruct} & $42.35$ & $56.00$ & $29.54$ & $35.30$ & $26.14$ & $\mathbf{47.08}$ & $14.37$ \\
\rowcolor{black!5}
\multicolumn{8}{c}{\strut\emph{External Memory}} \\
\hspace{.45em}+ BM25 RAG & $40.35_{\scriptstyle -2.00}$ & $52.83_{\scriptstyle -3.17}$ & $24.43_{\scriptstyle -5.11}$ & $32.20_{\scriptstyle -3.10}$ & $9.74_{\scriptstyle -16.40}$ & $37.86_{\scriptstyle -9.22}$ & $15.00_{\scriptstyle +0.63}$ \\
\hspace{.45em}+ LLMLingua-2 & $36.93_{\scriptstyle -5.42}$ & $50.03_{\scriptstyle -5.97}$ & $15.63_{\scriptstyle -13.91}$ & $21.45_{\scriptstyle -13.85}$ & $1.43_{\scriptstyle -24.71}$ & $38.45_{\scriptstyle -8.63}$ & $8.62_{\scriptstyle -5.75}$ \\
\hspace{.45em}+ MemoryBank & \textemdash & \textemdash & $17.65_{\scriptstyle -11.89}$ & $22.65_{\scriptstyle -12.65}$ & $7.67_{\scriptstyle -18.47}$ & $36.36_{\scriptstyle -10.72}$ & $9.88_{\scriptstyle -4.49}$ \\
\rowcolor{black!5}
\multicolumn{8}{c}{\strut\emph{Parametric Memory}} \\
\hspace{.45em}+ Context2LoRA & $37.85_{\scriptstyle -4.50}$ & $50.88_{\scriptstyle -5.12}$ & $32.53_{\scriptstyle +2.99}$ & $40.00_{\scriptstyle +4.70}$ & $29.86_{\scriptstyle +3.72}$ & $25.15_{\scriptstyle -21.93}$ & $17.75_{\scriptstyle +3.38}$ \\
\hspace{.45em}+ MemGen & $5.36_{\scriptstyle -36.99}$ & $16.27_{\scriptstyle -39.73}$ & $29.61_{\scriptstyle +0.07}$ & $34.85_{\scriptstyle -0.45}$ & $28.45_{\scriptstyle +2.31}$ & $44.30_{\scriptstyle -2.78}$ & $14.38_{\scriptstyle +0.01}$ \\
\rowcolor{black!5}
\multicolumn{8}{c}{\strut\emph{Auxiliary Memory}} \\
\hspace{.45em}+ MLP Memory & $10.94_{\scriptstyle -31.41}$ & $25.83_{\scriptstyle -30.17}$ & $28.80_{\scriptstyle -0.74}$ & $35.35_{\scriptstyle +0.05}$ & $26.00_{\scriptstyle -0.14}$ & $31.19_{\scriptstyle -15.89}$ & $14.38_{\scriptstyle +0.01}$ \\
\rowcolor{black!5}
\multicolumn{8}{c}{\strut\emph{Online Memory}} \\
\hspace{.45em}+ $\delta$-Mem (MSW) & $46.86_{\scriptstyle +4.51}$ & $60.47_{\scriptstyle +4.47}$ & $38.85_{\scriptstyle +9.31}$ & $44.40_{\scriptstyle +9.10}$ & $47.29_{\scriptstyle +21.15}$ & $41.55_{\scriptstyle -5.53}$ & $17.00_{\scriptstyle +2.63}$ \\
\cmidrule(lr){1-8}
\hspace{.45em}+ \textbf{LSTMem (Ours)} & $\mathbf{53.76}_{\scriptstyle +11.41}$ & $\mathbf{68.12}_{\scriptstyle +12.12}$ & $\mathbf{45.08}_{\scriptstyle +15.54}$ & $\mathbf{55.70}_{\scriptstyle +20.40}$ & $\mathbf{47.50}_{\scriptstyle +21.36}$ & $37.59_{\scriptstyle -9.49}$ & $\mathbf{18.00}_{\scriptstyle +3.63}$ \\
\midrule
\textbf{Qwen3-8B} & $32.48$ & $41.42$ & $31.87$ & $45.10$ & $12.79$ & $\mathbf{48.22}$ & $12.00$ \\
\rowcolor{black!5}
\multicolumn{8}{c}{\strut\emph{Parametric Memory}} \\
\hspace{.45em}+ Context2LoRA & $36.22_{\scriptstyle +3.74}$ & $49.19_{\scriptstyle +7.77}$ & $30.53_{\scriptstyle -1.34}$ & $43.15_{\scriptstyle -1.95}$ & $10.05_{\scriptstyle -2.74}$ & $34.04_{\scriptstyle -14.18}$ & $16.13_{\scriptstyle +4.13}$ \\
\rowcolor{black!5}
\multicolumn{8}{c}{\strut\emph{Online Memory}} \\
\hspace{.45em}+ $\delta$-Mem (MSW) & $40.15_{\scriptstyle +7.67}$ & $51.34_{\scriptstyle +9.92}$ & $32.66_{\scriptstyle +0.79}$ & $45.55_{\scriptstyle +0.45}$ & $14.95_{\scriptstyle +2.16}$ & $44.60_{\scriptstyle -3.62}$ & $13.38_{\scriptstyle +1.38}$ \\
\cmidrule(lr){1-8}
\hspace{.45em}+ \textbf{LSTMem (Ours)} & $\mathbf{53.76}_{\scriptstyle +21.28}$ & $\mathbf{67.49}_{\scriptstyle +26.07}$ & $\mathbf{34.61}_{\scriptstyle +2.74}$ & $\mathbf{46.80}_{\scriptstyle +1.70}$ & $\mathbf{18.74}_{\scriptstyle +5.95}$ & $41.70_{\scriptstyle -6.52}$ & $\mathbf{16.50}_{\scriptstyle +4.50}$ \\
\midrule
\textbf{SmolLM3-3B} & $1.67$ & $14.40$ & $11.08$ & $12.57$ & $5.53$ & $30.72$ & $8.00$ \\
\rowcolor{black!5}
\multicolumn{8}{c}{\strut\emph{Parametric Memory}} \\
\hspace{.45em}+ Context2LoRA & $30.28_{\scriptstyle +28.61}$ & $44.39_{\scriptstyle +29.99}$ & $14.23_{\scriptstyle +3.15}$ & $16.08_{\scriptstyle +3.51}$ & $2.86_{\scriptstyle -2.67}$ & $36.77_{\scriptstyle +6.05}$ & $14.75_{\scriptstyle +6.75}$ \\
\rowcolor{black!5}
\multicolumn{8}{c}{\strut\emph{Online Memory}} \\
\hspace{.45em}+ $\delta$-Mem (MSW) & $31.61_{\scriptstyle +29.94}$ & $46.77_{\scriptstyle +32.37}$ & $16.81_{\scriptstyle +5.73}$ & $18.10_{\scriptstyle +5.53}$ & $8.32_{\scriptstyle +2.79}$ & $\mathbf{39.63}_{\scriptstyle +8.91}$ & $\mathbf{16.12}_{\scriptstyle +8.12}$ \\
\cmidrule(lr){1-8}
\hspace{.45em}+ \textbf{LSTMem (Ours)} & $\mathbf{46.55}_{\scriptstyle +44.88}$ & $\mathbf{59.65}_{\scriptstyle +45.25}$ & $\mathbf{19.89}_{\scriptstyle +8.81}$ & $\mathbf{22.10}_{\scriptstyle +9.53}$ & $\mathbf{14.33}_{\scriptstyle +8.80}$ & $37.98_{\scriptstyle +7.26}$ & $15.38_{\scriptstyle +7.38}$ \\
\bottomrule
\end{tabular}
}
\endgroup

\end{table}

With Qwen3-4B-Instruct as the base model, \lstmem{} achieves the highest average scores on MAB and LoCoMo among the methods compared in Tables~\ref{tab:main} and~\ref{tab:locomo_cat}: $45.08$ on MAB and $53.60$ F1 on LoCoMo, $6.23$ and $4.48$ points above the published \dmem{} MSW results.
The gains concentrate on tasks that require retaining specific information from a long history and drawing on it later, the capability that motivates \lstmem{}.
Accurate retrieval contributes $6.16$ of the $6.23$-point MAB gain over \dmem{} ($+11.30$), and on all three backbones \lstmem{} is higher than \dmem{} MSW in both accurate retrieval and test-time learning; on Qwen3-4B-Instruct, selective forgetting is also $1.00$ point higher.
The flip side is LRU, which requires a holistic reading of a long narrative rather than retaining specific details, hence, on Qwen3-4B-Instruct and Qwen3-8B, every memory-augmented method in Table~\ref{tab:main} falls below the unadapted model on this family.
LoCoMo most directly reflects the long-horizon setting behind our design, since the memory is carried across the sessions of each conversation; there, \lstmem{} obtains the best score among all methods in every question category, with the largest gain over \dmem{} on single-hop questions ($+6.16$).
HotpotQA, which requires locating specific facts in several passages and combining them, gives the largest benchmark-level margin over \dmem{} on every backbone: \lstmem{} reaches $68.12$ F1 on Qwen3-4B, $7.65$ points above \dmem{}, and the margin widens to $16.15$ and $12.88$ points on Qwen3-8B and SmolLM3-3B.

\begin{table}[t]
\centering
\begingroup
\makeatletter
\makeatother
\begin{minipage}[c]{0.575\linewidth}
\caption{\textbf{LoCoMo F1 (\%).} Avg.\ is weighted over 1,540 questions.
Multi, Temp., Open and Single denote multi-hop, temporal, open-domain and single-hop questions.}
\label{tab:locomo_cat}
\vspace{6pt}
\begingroup
\fontsize{9}{11}\selectfont
\renewcommand{\arraystretch}{1.14}
\setlength{\heavyrulewidth}{.65pt}
\setlength{\lightrulewidth}{.35pt}
\setlength{\tabcolsep}{3pt}
\begin{tabular*}{\linewidth}{@{\extracolsep{\fill}}lrrrrr@{}}
\toprule
\textbf{Model} & \textbf{Avg.} & \textbf{Multi} & \textbf{Temp.} & \textbf{Open} & \textbf{Single} \\
\midrule
Qwen3-4B-Instruct & 40.79 & 38.39 & 32.89 & 10.77 & 48.05 \\
\hspace{.45em}+ BM25 RAG & 36.68 & 38.12 & 20.34 & 9.99 & 45.47 \\
\hspace{.45em}+ LLMLingua-2 & 40.98 & 39.07 & 30.13 & 10.98 & 49.19 \\
\hspace{.45em}+ MemoryBank & 38.14 & 37.88 & 21.76 & 13.35 & 47.31 \\
\hspace{.45em}+ Context2LoRA & 48.11 & 37.95 & 34.99 & 16.75 & 60.11 \\
\hspace{.45em}+ MemGen & 40.05 & 32.93 & 33.30 & 12.67 & 48.13 \\
\hspace{.45em}+ MLP Memory & 26.85 & 32.87 & 16.72 & 8.81 & 30.75 \\
\hspace{.45em}+ $\delta$-Mem (MSW) & 49.12 & 42.57 & 39.31 & 18.12 & 58.59 \\
\midrule
\hspace{.45em}+ \textbf{LSTMem (Ours)} & \textbf{53.60} & \textbf{46.25} & \textbf{40.94} & \textbf{19.92} & \textbf{64.75} \\
\bottomrule
\end{tabular*}
\par
\endgroup

\end{minipage}\hfill%
\begin{minipage}[c]{0.40\linewidth}
\caption{\textbf{General-capability scores (\%).} IFEval uses strict prompt-level accuracy;
GPQA-D uses accuracy.}
\label{tab:general_capabilities}
\vspace{6pt}
\begingroup
\fontsize{9}{11}\selectfont
\renewcommand{\arraystretch}{1.14}
\setlength{\heavyrulewidth}{.65pt}
\setlength{\lightrulewidth}{.35pt}
\setlength{\tabcolsep}{1.5pt}
\begin{tabular*}{\linewidth}{@{\extracolsep{\fill}}lrr@{}}
\toprule
\textbf{Model} & \textbf{IFEval} $\uparrow$ & \textbf{GPQA-D} $\uparrow$ \\
\midrule
Qwen3-4B-Instruct & 81.89 & 39.39 \\
\hspace{.45em}+ Context2LoRA & 76.71 & 29.29 \\
\hspace{.45em}+ MemGen & 39.37 & 38.89 \\
\hspace{.45em}+ MLP Memory & 24.95 & 22.73 \\
\hspace{.45em}+ $\delta$-Mem (MSW) & 81.52 & 37.37 \\
\midrule
\hspace{.45em}+ \textbf{LSTMem (Ours)} & 81.70 & 42.42 \\
\bottomrule
\end{tabular*}
\par
\endgroup

\end{minipage}
\par\vspace{6pt}

\endgroup
\end{table}

\lstmem{} obtains the best average score on every memory benchmark.
A memory interface should add history-dependent abilities while minimizing its impact on general ones \citep{biderman2024lora}, so we also evaluate general capabilities on Qwen3-4B-Instruct with a fresh memory for each example (Table~\ref{tab:general_capabilities}).
Relative to the unadapted model, \lstmem{} lowers IFEval strict prompt-level accuracy by only $0.19$ points and raises GPQA-Diamond accuracy by $3.03$ points, the best scores among all memory methods. Because the backbone remains frozen, these changes come only from the memory interface, and detaching it recovers the original model exactly.

\subsection{Ablation studies and analysis}
\label{sec:ablation}
\label{sec:c2}
\begin{table}[t]
\vspace{-6mm}
\centering
\begin{minipage}[c]{0.49\linewidth}
\centering
\caption{\textbf{Depth ablation.} $\Delta$ = on $-$ off. The coupling and feedback are removed in training and inference alike.}
\label{tab:c2}
\begingroup
\scriptsize
\setlength{\tabcolsep}{2pt}
\renewcommand{\arraystretch}{1.1}
\resizebox{\linewidth}{!}{%
\begin{tabular}{@{}llrrrrr@{}}
\toprule
Seed & \shortstack{Depth\\coupling} & \shortstack{MAB\\Avg.} & AR & TTL & LRU & SF \\
\midrule
\multirow{3}{*}{42} & on  & 45.08 & 55.70 & 47.50 & 37.59 & 18.00 \\
 & off & 43.16 & 54.30 & 41.29 & 38.21 & 18.00 \\
 & $\Delta$ & $+1.92$ & $+1.40$ & $+6.21$ & $-0.62$ & $0.00$ \\
\midrule
\multirow{3}{*}{43} & on  & 45.45 & 54.90 & 52.36 & 37.44 & 17.50 \\
 & off & 43.68 & 54.35 & 44.36 & 37.39 & 17.75 \\
 & $\Delta$ & $+1.77$ & $+0.55$ & $+8.00$ & $+0.05$ & $-0.25$ \\
\bottomrule
\end{tabular}%
}
\endgroup
\end{minipage}\hfill
\begin{minipage}[c]{0.49\linewidth}
\centering
\caption{\textbf{Memory placement ablation on LoCoMo.} F1 (\%) on Qwen3-4B-Instruct. Layer indices are one-based: even/odd select 18 alternating layers; first/middle/last select layers 1--12, 13--24 and 25--36, respectively.}
\label{tab:layer_placement_locomo}
\begingroup
\scriptsize
\setlength{\tabcolsep}{2pt}
\renewcommand{\arraystretch}{1.1}
\resizebox{\linewidth}{!}{%
\begin{tabular}{@{}lrrrrr@{}}
\toprule
Memory layout & Avg. & Multi & Temp. & Open & Single \\
\midrule
All 36  & 53.60 & 46.25 & 40.94 & 19.92 & 64.75 \\
Even 18 & 51.30 & 45.97 & 40.51 & 16.10 & 61.22 \\
Odd 18 & 51.56 & 46.83 & 39.73 & 14.10 & 61.93 \\
First 12 & 48.09 & 43.91 & 33.98 & 11.12 & 59.10 \\
Middle 12 & 50.42 & 46.93 & 39.19 & 13.94 & 60.03 \\
Last 12 & 50.58 & 47.50 & 41.37 & 14.28 & 59.28 \\
\bottomrule
\end{tabular}%
}
\endgroup

\end{minipage}
\vspace{-2mm}
\end{table}

\textbf{Ablation on depth-state connection.}
In \lstmem{}, each memory layer predicts its write from the hidden memory $D_t^{(\ell-1)}$ of the layer below and carries that memory into its own expressed state (Section~\ref{sec:depth}).
Replacing $D_t^{(\ell-1)}$ with zeros in both training and inference lowers MAB by $1.92$ and $1.77$ points (Table~\ref{tab:c2}), both larger than the $1.2$-point difference between two retrainings of the coupled model with the same seed.
The drop is concentrated in test-time learning ($6.21$ and $8.00$ points); accurate retrieval also decreases on both seeds, while LRU and SF change by at most $0.62$ points.
Test-time learning requires learning new tasks from examples accumulated in the history rather than recalling a single fact.
This matches the motivation for the hierarchy: with depth coupling, each layer writes only what the layer below does not already predict, so higher layers can build on lower-layer memory instead of duplicating it.

\textbf{Ablation on different layers.}
We retrain the main configuration with memory on subsets of the 36 layers (Table~\ref{tab:layer_placement_locomo}) and take $D_t^{(\ell-1)}$ from the preceding memory-enabled layer.
The full stack exceeds the best sparse layout by $2.04$ points, and the average grows with the number of memory layers: $48.09$--$50.58$ with 12 layers, $51.30$--$51.56$ with 18 and $53.60$ with 36.
No sparse layout exceeds the full stack by more than $1.25$ points in any individual question category, whereas every sparse layout trails it by at least $3.82$ points on open-domain and $2.82$ points on single-hop questions.
These results support extending the memory hierarchy through the full depth of the backbone.

\textbf{Ablation on training stages.}
Our main model continues training on the Long dataset after QASPER, whereas the published \dmem{} results use QASPER alone, so we also train \dmem{} with the same second stage (Table~\ref{tab:c1}).
This stage does improve \dmem{}, raising its LoCoMo average to $50.06$.
It helps \lstmem{} more: \lstmem{} gains $3.42$ points, improves in every question category, and reaches $53.60$, $3.54$ points above \dmem{} after the same two stages.
Even after both stages, \dmem{} remains below \lstmem{} after Stage~1 alone ($50.18$).
This indicates that the advantage of \lstmem{} stems from the architecture, and the gap widens as training continues on long histories.

\textbf{Ablation on block-end feedback.}
Within a history block, memory flows only from shallow to deep layers; block-end feedback adds the reverse direction, letting lower layers correct their cells with the reconstruction errors of higher layers before the next block (Section~\ref{sec:block_feedback}).
Disabling it while keeping forward depth coupling lowers the LoCoMo average from $53.60$ to $51.89$ (Table~\ref{tab:feedback}).
The loss falls on single-hop and open-domain questions ($2.82$ and $2.44$ points), whereas multi-hop and temporal scores change by less than $0.2$ points.
Feedback thus accounts for $2.82$ of the $6.16$-point single-hop advantage of \lstmem{} over \dmem{} in Table~\ref{tab:locomo_cat}, suggesting that it mainly sharpens the recall of individual facts stated in the conversation.
Even without feedback, \lstmem{} remains above \dmem{} trained with the same two stages ($51.89$ vs.\ $50.06$; Table~\ref{tab:c1}), and feedback widens this margin to $3.54$ points.
The feedback correction is computed from key--value associations already in the history, without answer labels or parameter updates.

\begin{table}[t]
\centering
\caption{\textbf{Training-stage comparison on LoCoMo.} Stage 1 trains on QASPER, and Stage 2 continues training on the Long dataset. Scores are F1 (\%), with Avg.\ weighted by question count.}
\label{tab:c1}
\begingroup
\small
\setlength{\tabcolsep}{4pt}
\renewcommand{\arraystretch}{1.1}
\resizebox{0.85\linewidth}{!}{%
\begin{tabular}{@{}llccccc@{}}
\toprule
Method & Training stage & Multi-hop & Temporal & Open-domain & Single-hop & Avg. \\
\midrule
\multirow{2}{*}{$\delta$-Mem} & 1: QASPER & 42.57 & 39.31 & 18.12 & 58.59 & \textbf{49.12} \\
 & 2: + Long & 46.73 & 38.75 & 16.20 & 59.36 & \textbf{50.06} \\
\midrule
\multirow{2}{*}{LSTMem} & 1: QASPER & 43.70 & 39.03 & 18.72 & 60.19 & \textbf{50.18} \\
 & 2: + Long & 46.25 & 40.94 & 19.92 & 64.75 & \textbf{53.60} \\
\bottomrule
\end{tabular}%
}
\endgroup
\vspace{-4mm}
\end{table}

\begin{table}[t]
\centering
\caption{\textbf{Block-end feedback ablation on LoCoMo.} F1 (\%) on Qwen3-4B-Instruct. Both rows keep forward depth coupling and differ only in whether block-end feedback is applied. }
\label{tab:feedback}
\begingroup
\small
\setlength{\tabcolsep}{7pt}
\begin{tabular}{@{}lrrrrr@{}}
\toprule
Block-end feedback & Avg. & Multi & Temp. & Open & Single \\
\midrule
On & 53.60 & 46.25 & 40.94 & 19.92 & 64.75 \\
Off & 51.89 & 46.31 & 40.78 & 17.48 & 61.93 \\
\bottomrule
\end{tabular}
\endgroup
\vspace{-3mm}
\end{table}

\section{Conclusion}
\label{sec:conclusion}

We introduced \lstmem{}, an online long short-term memory that complements a language model's current context with a persistent stack of gated memory cells. The cell state accumulates history, the hidden state expresses memory for current use, and hidden-state propagation conditions upper memories on lower ones; block-end feedback closes the loop by letting lower layers correct their cells with the reconstruction errors of higher layers, without answer labels or parameter updates. With the backbone frozen, \lstmem{} obtains the best MemoryAgentBench averages and HotpotQA scores among all compared methods on all three backbones, as well as the best LoCoMo score on Qwen3-4B-Instruct, with gains concentrated on tasks that require retaining specific information from a long history. Ablations support each design choice: removing depth coupling lowers MemoryAgentBench on both training seeds, memory on all layers gives the best LoCoMo average, disabling feedback lowers LoCoMo, and the advantage over \dmem{} persists when both are trained with the same stages. These findings motivate treating online memory as both a temporal process and a hierarchy across model depth.

\section*{AI Use Statement}
In this work, we used generative AI tools for [generating synthetic data sets, helping develop theoretical models or conceptual frameworks, designing research methods or experiments or providing feedback on them, implementing methods, assisting with translation].
We have not used generative AI tools for [supporting qualitative and thematic data analysis, interpreting results], and [formulating mathematical claims, providing key elements for proving mathematical claims, assisting in writing proofs, generating or refining hypotheses, cleaning and reformatting data sets] are not applicable to this work.
We have reviewed all AI-assisted work. In particular, all LLM-generated code was verified and tested for correctness by two authors.
We take responsibility for the final content of this work, including text, claims or artifacts produced with the aid of generative AI.
\section*{Ethics statement}
\label{sec:ethics}
This work studies online memory for an open-weight language model using public benchmark data and constructed training streams. We do not recruit human participants or collect new user interactions. A deployed persistent memory may retain information beyond a single session, so users should be able to control memory updates and reset the state.

\section*{Reproducibility statement}
Section~\ref{sec:method} and Appendix~\ref{app:memory_impl} specify the memory updates, block-end feedback, and training recipe. Appendix~\ref{app:protocol} gives the evaluation protocol, and Appendix~\ref{app:efficiency} the efficiency setup. We release our code as well.

\clearpage
\appendix
\addtocontents{toc}{\protect\setcounter{tocdepth}{2}}
\begingroup
\renewcommand{\contentsname}{Appendix Contents}
\tableofcontents
\endgroup
\section{Related Work}
\label{sec:related}

\paragraph{Online Memory.}
Online memory incorporates new information by updating its contents during inference \citep{zhoubian2026memorylargelanguagemodels,sun2025ttt,behrouz2025titans,behrouz2025atlas,wang2025testtime,packer2023memgpt}.
Linear attention admits a recurrent matrix-state formulation \citep{katharopoulos2020linear,sun2023retnet,dao2024transformers,peng2024eagle}, connecting sequence processing to fast-weight memories that store key--value associations \citep{schlag2021linear}.
Recurrent matrix memories implement online updates by incorporating incoming associations into this fixed-size state \citep{gu2024mamba,qin2024hgrn2,liu2025longhorn,kimiteam2025kimilinear}.
Gated Linear Attention learns input-dependent retention gates \citep{yang2024gla}, while DeltaNet corrects stored associations using value-prediction errors \citep{yang2024deltanet}.
Gated DeltaNet combines these mechanisms to control forgetting and targeted writes \citep{yang2025gated}.
\dmem{} connects associative memory to pretrained language models by converting state readouts into low-rank attention corrections \citep{lei2026deltamem,lei2026transmemtransforminghiddenstates}.
Although their update rules differ, these methods read from the matrix in which they accumulate historical associations.
\lstmem{} separates these roles with a cell state $C_t$ for historical accumulation and a hidden state $D_t$ for reading, with input, forget, and output gates controlling how memory is written, retained, and expressed.

\paragraph{Long-term Memory.}
Long-term memory aims to keep earlier information available beyond the current context window \citep{zhoubian2026memorylargelanguagemodels,zhang2025agentsurvey,du2025rethinking}.
Textual memory systems such as MemoryBank retain interaction records for later retrieval \citep{zhong2024memorybank,packer2023memgpt,park2023generative,chhikara2025mem0,xu2025amem}.
LongMem instead caches past key--value representations and retrieves them through a residual side network \citep{wang2023longmem}.
Compressing history into a fixed-size memory provides another way to extend this horizon \citep{chevalier2023autocompressors,ge2024icae,munkhdalai2024infini,yu2026memagent}: MEMORYLLM maintains an updatable latent memory pool \citep{wang2024memoryllm,wang2025mplus}, while xLSTM uses gated recurrent memories, including a matrix-valued cell in mLSTM \citep{beck2024xlstm}.
Other approaches encode knowledge in LoRA modules \citep{hu2022lora,back2026context2lora,wang2024templora,chen2025generativeadapter} or generate latent memory tokens during reasoning, as in MemGen \citep{zhang2025memgen}.
Titans combines local attention with a neural memory updated at test time, using gradient-based surprise to guide memorization \citep{behrouz2025titans}.
\lstmem{} places recurrent memory alongside a pretrained Transformer and couples these memories across depth.
At each token position, the hidden matrix $D_t^{(\ell-1)}$ passes to the next memory module, where it conditions both state construction and readout.

\section{Layer-wise Knowledge Editing as a Memory Probe}
\label{app:layer_editing}

This appendix details the layer-wise knowledge editing study behind the depth argument in Section~\ref{sec:intro}. We use knowledge editing as a controlled probe of memory in a language model. Editing a fact writes a new memory into the model, and the three standard editing metrics then describe how this memory behaves. Efficacy measures whether the written memory can be recalled from the cue it was written with, generalization measures whether it can be recalled from a rephrased cue, and specificity measures whether an unrelated existing memory is retained after the write. Writing the same memories into different bands of layers under a fixed parameter budget therefore tests whether storing and retrieving memory depends on depth.

\paragraph{Setup.}
We edit Qwen3.5-2B on ZsRE \citep{levy2017zsre} with LoRA \citep{hu2022lora} of rank 8 and alpha 16. Qwen3.5-2B has 24 layers, with full attention in every fourth layer and linear attention in the others. The update is restricted to one of three contiguous bands of eight layers, the shallow band (layers 1--8), the middle band (layers 9--16), or the deep band (layers 17--24). Within the band, LoRA is applied to the query, key, value, and output projections of the full-attention layers and to the query-key-value input projection and the output projection of the linear-attention layers, so every layer in the band is updated. Each band contains two full-attention and six linear-attention layers and therefore the same 835,584 trainable LoRA parameters, and all bands share the same optimization protocol, so the comparison varies only in the depth at which the memory is written.

All 19,086 evaluation records are edited sequentially in shards of 512 consecutive records. Each shard starts from the unedited model, and a single LoRA adapter and AdamW optimizer persist across its records. Each record is written with eight optimization steps at a learning rate of \mbox{$5\times10^{-4}$} on a chat-format question-answering prompt and is evaluated immediately after its update, so later records are edited on top of earlier ones. The loss is the cross-entropy on the answer tokens plus $0.75$ times a locality term, the negative log-likelihood of the reference answer to the record's neighborhood prompt, which is also the prompt used to measure specificity.

Evaluation follows the token-level protocol of \citet{meng2022rome}. For efficacy and generalization, the target answer is appended to the edited prompt or to its paraphrase, and each answer token counts as correct when it is the model's top prediction given the preceding tokens. Specificity uses the record's neighborhood prompt, which asks about an unrelated fact, and counts it as correct when the model's top next-token prediction is the first token of the reference answer. Accuracies are pooled over tokens across records, and the editing score is the harmonic mean of the three. All scores are percentages, and higher is better.

Sequential editing occasionally diverges, after which the adapter no longer fits new records and the model produces degenerate text. We flag a shard as diverged when at least 50 of its records, or 15 of its last 20, end training with a loss above 5. This occurred in 5 of the 114 shard runs across the three bands (one shallow, one middle, and three deep), and each converged when rerun once with a different random seed. Table~\ref{tab:layer_editing_zsre} uses these reruns. Keeping the diverged shards instead gives editing scores of 62.73, 68.95, and 65.97 for the shallow, middle, and deep bands, which leaves their ordering unchanged.

\begin{table}[ht]
\centering
\caption{\textbf{Writing memories into one layer band of Qwen3.5-2B (knowledge editing on ZsRE).} Scores (\%) on all 19,086 records. Efficacy and generalization measure recall of the written memory from its original and a rephrased cue, and specificity measures retention of an unrelated memory. Layers are numbered from the input, and bold marks the best band in each column.}
\label{tab:layer_editing_zsre}
\small
\begin{tabular}{@{}lrrrr@{}}
\toprule
Layer band & Efficacy & Generalization & Specificity & Editing score \\
\midrule
Shallow (1--8) & 56.89 & 51.57 & \textbf{94.40} & 63.08 \\
Middle (9--16) & 63.15 & 60.06 & 92.70 & \textbf{69.33} \\
Deep (17--24) & \textbf{64.81} & \textbf{63.64} & 74.60 & 67.34 \\
\bottomrule
\end{tabular}
\end{table}

\paragraph{Results.}
The depth at which a memory is written strongly shapes how it behaves (Table~\ref{tab:layer_editing_zsre}). Writing into the middle band rather than the shallow band improves recall of the new memory by $6.26$ points from its original cue and by $8.49$ points from a rephrased cue, at a cost of only $1.70$ points in retaining unrelated memories. Writing into the deep band adds just $1.66$ and $3.58$ points of recall but lowers retention by $18.10$ points, even though the locality term directly protects the evaluated memory. From shallow to deep layers, recall of new memories thus improves while interference with existing memories grows sharply, and the balance between the two, summarized by the editing score, peaks at the middle band. Although this probe modifies model weights rather than online memory states, it reveals that knowledge-editing outcomes are sensitive to the depth of the updated layers under a fixed parameter budget and optimization protocol.
This depth sensitivity motivates us to investigate how online memory should be organized across model depth.
We explore cross-layer memory propagation in \lstmem{} (Section~\ref{sec:method}) and evaluate its contribution through a separate depth ablation (Section~\ref{sec:c2}).

\section{Evaluation protocol}
\label{app:protocol}
\label{app:metrology}

\paragraph{MemoryAgentBench.}
For MemoryAgentBench \citep{hu2025memoryagentbench}, all 3,671 questions are scored with the per-source metrics: normalised alias accuracy for question answering, F1 for InfiniteBench summarisation \citep{zhang2024infinitebench}, and Recall@5 for movie recommendation. Family and overall scores are weighted by question count. The generation limits are 40 tokens for EventQA, 50 for LongMemEval \citep{wu2025longmemeval} and RULER \citep{hsieh2024ruler}, 10 for FactConsolidation, 20 for in-context classification, 2,000 for Detective QA \citep{xu2024detectiveqa}, 1,200 for summarisation and 300 for movie recommendation. The prompt requests a short answer supported by the supplied context and an explicit abstention when unsupported.

\paragraph{LoCoMo.}
For LoCoMo \citep{maharana2024locomo}, we use categories 1--4: multi-hop, temporal, open-domain and single-hop, totalling 1,540 questions. Category 5 is excluded as in \citet{lei2026deltamem}. Conversations are replayed session by session through the memory, which is reset before each independent conversation. Questions use the official single-prompt format and a 50-token answer limit. Scores use token-level F1 with Porter stemming and the official multi-answer handling; the total is a question-weighted mean of category scores.

\paragraph{HotpotQA, IFEval and GPQA-Diamond.}
HotpotQA \citep{yang2018hotpotqa} uses 7,405 validation questions with supporting passages, a 32-token generation limit and EM/F1 scoring. IFEval \citep{zhou2023ifeval} uses 541 prompts, at most 1,500 generated tokens and strict prompt-level accuracy. GPQA-Diamond \citep{rein2023gpqa} uses 198 questions, a JSON answer-letter prompt and at most 8,192 generated tokens. All five benchmarks use sampled decoding with temperature 0.4, top-$p$ 0.9, top-$k$ 10 and seed 42. General-capability evaluations initialise a fresh memory for each independent example.

\textbf{Depth-ablation settings.}
The depth ablation trains and evaluates both settings with the frozen backbone. We report individual training seeds and do not estimate benchmark confidence intervals. The two historical retrainings mentioned in Section~\ref{sec:c2} provide a variability reference, not a statistical bound.

\section{Memory implementation details}
\label{app:memory_impl}

\subsection{State configuration and attention interface}

Our main Qwen3-4B-Instruct configuration attaches memory modules to all 36 Transformer layers, with $n_h=4$ heads and $r=8$.
All backbone parameters remain frozen throughout both training stages, and only the memory adapter is trained.
The denominator in the query/key normalization is bounded below by $10^{-6}$.
Both temporal and depth reads use direct matrix--vector products without additional read-side selection or decay.
We use $\alpha=16$, unit extra output scale and no memory-use gate.
Native backbone processing is preserved, including Qwen3 key normalization \citep{yang2025qwen3} and SmolLM3's layer-specific RoPE/no-RoPE choice \citep{bakouch2025smollm3,su2024roformer,yang2025ropenope}.

\subsection{Effective projection blocks}
\label{app:projection_blocks}

The prediction and gate projections in the full parameterization receive a zero block in place of the current layer's temporal-memory input.
Partition their weight matrices according to these inputs:
\begin{equation*}
W_{\mathrm{pred}}^{\mathrm{full}}=[P_{\mathrm{time}}\mid P_{\mathrm{dep}}],\qquad
W_{\mathrm{gate}}^{\mathrm{full}}=[G_z\mid G_{\mathrm{time}}\mid G_{\mathrm{dep}}\mid G_e].
\end{equation*}
The effective matrices used in Eqs.~\ref{eq:residual} and~\ref{eq:gates} are
\begin{equation*}
\widetilde W_{\mathrm{pred}}=P_{\mathrm{dep}},\qquad
\widetilde W_{\mathrm{gate}}=[G_z\mid G_{\mathrm{dep}}\mid G_e].
\end{equation*}
They give the same forward maps:
\begin{equation*}
\begin{aligned}
W_{\mathrm{pred}}^{\mathrm{full}}[0;p_t^{\mathrm{dep}}]
&=\widetilde W_{\mathrm{pred}}p_t^{\mathrm{dep}},\\
W_{\mathrm{gate}}^{\mathrm{full}}[z_t;0;r_t^{\mathrm{dep}};e_t]
&=\widetilde W_{\mathrm{gate}}[z_t;r_t^{\mathrm{dep}};e_t].
\end{aligned}
\end{equation*}
Biases are unchanged.
The full matrices are retained in the stored parameterization; the main-text matrices select their active blocks.
Neither projection takes the current layer's temporal memory as an explicit input.
In particular, initializing $W_{\mathrm{pred}}^{\mathrm{full}}=[I_{d_s}\mid0_{d_s\times d_s}]$ corresponds to $\widetilde W_{\mathrm{pred}}=0$.

All memory readouts, residuals, and within-layer features $r_t^{\mathrm{time}},r_t^{\mathrm{dep}},p_t^{\mathrm{dep}},\hat v_t,e_t,u_t,z_t$ have width $d_s$.
Let $d_q$ denote the output width of the backbone query projection.
The projection dimensions are
\begin{equation*}
\begin{aligned}
W_q^{\mathrm{mem}},W_k^{\mathrm{mem}},W_v^{\mathrm{mem}},W_x
&\in\mathbb{R}^{d_s\times d},\\
W_{\mathrm{read}},W_{\mathrm{pred}}^{\mathrm{full}}
&\in\mathbb{R}^{d_s\times2d_s},\\
\widetilde W_{\mathrm{pred}}&\in\mathbb{R}^{d_s\times d_s},\\
\widetilde W_{\mathrm{gate}}&\in\mathbb{R}^{4d_s\times3d_s},\\
W_{\mathrm{gate}}^{\mathrm{full}}&\in\mathbb{R}^{4d_s\times4d_s},\\
W_{\Delta q}&\in\mathbb{R}^{d_q\times d_s},\\
W_{\Delta o}&\in\mathbb{R}^{d\times d_s}.
\end{aligned}
\end{equation*}
The biases satisfy $b_x,b_{\mathrm{read}},b_{\mathrm{pred}}\in\mathbb{R}^{d_s}$ and $b_{\mathrm{gate}}\in\mathbb{R}^{4d_s}$.

\subsection{Parallel state updates}
\label{app:parallel_scan}

Once the layer inputs and lower-layer hidden sequence are available, the candidates and gates can be computed independently of the current layer's cell recurrence.
Fixing a memory layer and omitting its layer index, the cell update for head $j$ in Eq.~\ref{eq:write} has the affine form
\begin{equation*}
\begin{aligned}
C_t^{[j]}&=A_t^{[j]}C_{t-1}^{[j]}+B_t^{[j]},\\
A_t^{[j]}&=\operatorname{diag}(f_t^{[j]}),\qquad
B_t^{[j]}=\operatorname{diag}(i_t^{[j]})e_t^{[j]}(k_t^{\mathrm{mem},[j]})^\top.
\end{aligned}
\end{equation*}
Affine updates compose associatively:
\begin{equation*}
(A_2,B_2)\circ(A_1,B_1)
=(A_2A_1,A_2B_1+B_2).
\end{equation*}
A parallel prefix scan \citep{blelloch1990prefix} computes the cumulative transformations, which are applied to the incoming cell state.
The hidden sequence then follows from Eq.~\ref{eq:hidden}.
We implement the scan in Triton \citep{tillet2019triton}.
Scanning is confined to a layer and a history block; block-end feedback determines the boundary state for the next block.

\subsection{Implementation of block-end feedback}
\label{app:feedback_impl}

For each history block containing valid positions, $\mathcal I_b$ contains at most 32 distinct, evenly spaced valid positions shared across layers.
For each sequence, let $T_b$ be the last valid token position in history block $b$.
Superscripts $\mathrm{bef}$ and $\mathrm{aft}$ denote states before and after feedback.
The pre-feedback states and cached gates are
\begin{equation*}
\begin{aligned}
(C_b^{(\ell),\mathrm{bef}},D_b^{(\ell),\mathrm{bef}})
&=(C_{T_b}^{(\ell)},D_{T_b}^{(\ell)}),\\
(o_b^{(\ell)},g_b^{(\ell)})
&=(o_{T_b}^{(\ell)},g_{T_b}^{(\ell)}),
\end{aligned}
\end{equation*}
where the states on the right are evaluated immediately before feedback.
Empty blocks are skipped.
All reconstruction losses and inner partial derivatives are evaluated at the same pre-feedback states.

In the following equations, a gate vector is reshaped to $n_h\times r\times1$ and broadcast over the last matrix dimension.
Thus, for a gate $g$ and a state-shaped tensor $A$,
\begin{equation*}
[g\odot A]_{j,a,c}=g_{j,a}A_{j,a,c}.
\end{equation*}
This is equivalent to left multiplication by $\operatorname{diag}(g^{[j]})$ within each head.
Both $\tanh$ and its square act elementwise, and $1$ in Eq.~\ref{eq:depth_feedback} denotes the all-ones tensor.

The compact gradient in Eq.~\ref{eq:feedback_step} follows only the explicit depth-state connection in Eq.~\ref{eq:hidden}.
For this derivative, sampled memory keys, values, and terminal gates are fixed.
This defines the inner state correction and does not detach these quantities from the outer training graph when history gradients are retained.
Let
$R_b^{(\ell)}=\partial\mathcal L_b^{(\ell)}/\partial D_b^{(\ell),\mathrm{bef}}$.
The contributions from all strictly higher layers can be accumulated as
\begin{equation}
\begin{aligned}
U_b^{(L)}&=0,\\
U_b^{(\ell)}
&=g_b^{(\ell+1)}\odot
\left(R_b^{(\ell+1)}+U_b^{(\ell+1)}\right),
\qquad \ell<L,\\
G_b^{(\ell)}
&=o_b^{(\ell)}\odot
\left(1-\tanh^2\!\left(C_b^{(\ell),\mathrm{bef}}\right)\right)
\odot U_b^{(\ell)}.
\end{aligned}
\label{eq:depth_feedback}
\end{equation}
Here $o_b^{(\ell)}$ and $g_b^{(\ell)}$ are the cached terminal gates.
This recurrence is the chain-rule expansion of the strictly higher-layer loss gradient in Eq.~\ref{eq:feedback_step}; it excludes the layer's own reconstruction loss.

The candidate correction is computed separately for each memory head, with $\epsilon=10^{-12}$ and $\eta=0.01$:
\begin{equation}
\begin{aligned}
a_b^{(\ell),[j]}
&=\operatorname{RMS}\!\left(G_b^{(\ell),[j]}\right),\\
s_b^{(\ell),[j]}
&=\max\!\left(1,\operatorname{RMS}\!\left(C_b^{(\ell),\mathrm{bef},[j]}\right)\right),\\
\widehat{\Delta C}_b^{(\ell),[j]}
&=\begin{cases}
-\eta s_b^{(\ell),[j]}G_b^{(\ell),[j]}/a_b^{(\ell),[j]},
&a_b^{(\ell),[j]}\ge\epsilon,\\
\mathbf{0},&a_b^{(\ell),[j]}<\epsilon.
\end{cases}
\end{aligned}
\label{eq:feedback_step_impl}
\end{equation}
RMS is computed over the $r\times r$ entries of a head, with correction arithmetic in FP32.
In the following acceptance rule, we suppress the block, layer, and head indices.
If the relative RMS of $\widehat{\Delta C}$ exceeds $\rho=0.05$, we proportionally rescale it to $\rho$.
We then form the candidate cell $C^{\mathrm{cand}}$ and cast it to the original state dtype.
The candidate is accepted only if
\begin{equation*}
\frac{\operatorname{RMS}\!\left(
\operatorname{fp32}(C^{\mathrm{cand}})-\operatorname{fp32}(C^{\mathrm{bef}})
\right)}{s}
\le\rho+10^{-7}.
\end{equation*}
Here $\operatorname{fp32}$ denotes conversion to FP32.
Otherwise, the original cell is retained.
Here $\Delta C=C^{\mathrm{aft}}-C^{\mathrm{bef}}$ denotes the accepted change, including dtype rounding.
Inactive heads retain their original states, and all accepted changes are committed simultaneously using Eq.~\ref{eq:feedback_step}.
Nonfinite feedback losses, gradients, or corrected states abort the feedback step.

Using the cached terminal gates, expressed states are then rebuilt from shallow to deep layers:
\begin{equation}
\begin{aligned}
D_b^{(\ell),\mathrm{aft}}
={}&D_b^{(\ell),\mathrm{bef}}
+o_b^{(\ell)}\odot
\left[
\tanh\!\left(C_b^{(\ell),\mathrm{aft}}\right)
-\tanh\!\left(C_b^{(\ell),\mathrm{bef}}\right)
\right]\\
&+g_b^{(\ell)}\odot
\left[
D_b^{(\ell-1),\mathrm{aft}}
-D_b^{(\ell-1),\mathrm{bef}}
\right].
\end{aligned}
\label{eq:feedback_hidden}
\end{equation}
For the first memory layer, we set $D_b^{(0),\mathrm{bef}}=D_b^{(0),\mathrm{aft}}=\mathbf{0}$, so the lower-layer increment is zero.
The incremental form preserves a state if neither its cell nor its lower-layer input changes.
The highest-layer cell is unchanged because $U_b^{(L)}=0$, while its expressed state may change through the corrected states below.
The next history block starts from $(C_b^{(\ell),\mathrm{aft}},D_b^{(\ell),\mathrm{aft}})$. Feedback leaves model parameters and outputs already computed in block $b$ unchanged.

\subsection{Training and initialization}
\label{app:training_call}

\paragraph{Episode and objective.}
Each training episode starts from $C_0^{(\ell)}=D_0^{(\ell)}=\mathbf{0}$ at every memory layer.
Available history is written into a memory stack $\mathcal M$, and the final assistant turn is predicted with writing and feedback disabled.
For $M$ supervised answer tokens $y_1,\ldots,y_M$, let $y_{<m}=(y_1,\ldots,y_{m-1})$. The objective is
\begin{equation}
\mathcal{L}_{\mathrm{ans}}=-\frac{1}{M}\sum_{m=1}^{M}
\log p_{\phi,\theta}(y_m\mid y_{<m},\mathrm{prompt},\mathcal{M}).
\label{eq:objective}
\end{equation}
Only adapter parameters $\theta$ are trained, while all backbone parameters $\phi$ remain frozen.
Answers start without historical KV cache, with a 512-token read-sequence limit and context-dropout probability 0.2.
Equation~\ref{eq:feedback_loss} defines the reconstruction objective used for the inner state correction. This objective is not added to the answer loss.

\paragraph{Long dataset.}
The second training stage uses the Long dataset, a synthetic collection of long-horizon memory episodes that we construct with a large language model.
Each history is generated through an API in successive chunks that are concatenated into one long context, and each context is paired with several questions whose answers depend on it.
We generate 100 candidate contexts in this way, using 1,350 chunk requests in total.
Automatic checks reject 29 contexts, for example because they contain email addresses outside example domains, leak conversation-role markers, or include a question that reveals its answer.
The remaining 71 contexts provide 832 question--answer pairs.
Generation targets histories of about 32K and 64K tokens, but the accepted histories run longer, with 36 contexts in the shorter group and 35 in the longer group, spanning 44,074 to 98,340 tokens under the Qwen3 tokenizer.
Each training example writes one full history into memory and then asks one of its questions, whose answer contains 6 to 15 tokens.
Because answering starts without the historical KV cache, the model must recover each answer from memory, so these episodes train it to read information retained over long histories.

\paragraph{History gradients.}
QASPER \citep{dasigi2021qasper} histories are capped at 8,192 tokens and retain gradients, allowing answer cross-entropy to differentiate through both memory updates and feedback correction.
Fixing keys, values and gates for the inner partial derivative does not detach them from this outer training graph.
Histories in the Long dataset are capped at 131,072 tokens, so none is truncated, and are written in 2,048-token blocks without gradients, with states detached between blocks.
Their answer passes train memory reading and attention correction.

\paragraph{Training schedule.}
We train a fresh adapter on 2,240 QASPER presentations for 70 updates at $2\times10^{-4}$, then on the 832 examples of the Long dataset for 26 updates at $5\times10^{-5}$ with a fresh optimizer.
Both stages use seed 42, batch size 32 (eight workers, microbatch one, accumulation four), AdamW \citep{loshchilov2019adamw}, BF16, ZeRO-2 \citep{rajbhandari2020zero}, 10\% warm-up and cosine decay.
Feedback from all higher layers is enabled at history-block boundaries during training and inference.
The feedback ablation disables feedback while retaining forward depth coupling, whereas the depth ablation sets the forward depth input $D_t^{(\ell-1)}=0$.
No auxiliary KL, teacher loss, write-sparsity penalty or additional forgetting stage is used.

\paragraph{Initialization.}
We initialize $W_{\mathrm{pred}}^{\mathrm{full}}$ and $W_{\mathrm{read}}$ separately to $[I_{d_s}\mid0_{d_s\times d_s}]$, with zero biases. In the effective parameterization, $\widetilde W_{\mathrm{pred}}=0$, so initially $e_t=v_t^{\mathrm{mem}}$ and read fusion selects temporal memory.
All gate weights, including $\widetilde W_{\mathrm{gate}}$, start at zero, with $i,f,o,g$ biases $-1,+2,0,-1$.
Memory projections and $W_x$ use Kaiming-uniform initialization, with $b_x=0$.
Correction projections use the first eight columns of the corresponding backbone query/output weights, normalized column-wise and scaled by 0.05, with remaining columns initialized to zero.

\paragraph{Online use.}
During inference, history blocks update and refine the carried $C/D$ states with model parameters fixed.
Memory readout continues during answer generation; answer-time state updates follow the corresponding evaluation protocol.
Evaluation prompts retain the history specified by their protocols, and memory is reset for independent conversations.
\paragraph{Memory state during evaluation.}
\label{sec:eva}
We evaluate \lstmem{} with the prompt templates and memory-state handling of \citet{lei2026deltamem} left unchanged, and our \dmem{} run after the second stage (Table~\ref{tab:c1}) uses the same evaluation code, so the two runs differ only in the memory module.
The memory is reset before every independent history, that is, before each MemoryAgentBench context, each LoCoMo conversation and each HotpotQA example, and every IFEval and GPQA-Diamond prompt starts from an empty memory.
The history is written in 2,048-token blocks, and block-end feedback follows every block, including the last one, before the question is processed.
The prompt also retains the history in the official format of each benchmark, so when answering, the backbone attends to the historical KV cache while the memory adds its attention corrections.
The question and the generated answer read the memory and update it token by token as in \dmem{}, whereas block-end feedback is applied only to history blocks.
All questions on the same history, including those of one LoCoMo conversation, start from the same memory state and historical KV cache obtained after the last history block, so updates made while answering one question never carry over to the next, and scores do not depend on question order.

\begin{figure}[t]
\centering
\begin{minipage}[t]{0.49\linewidth}
\centering
\includegraphics[width=\linewidth]{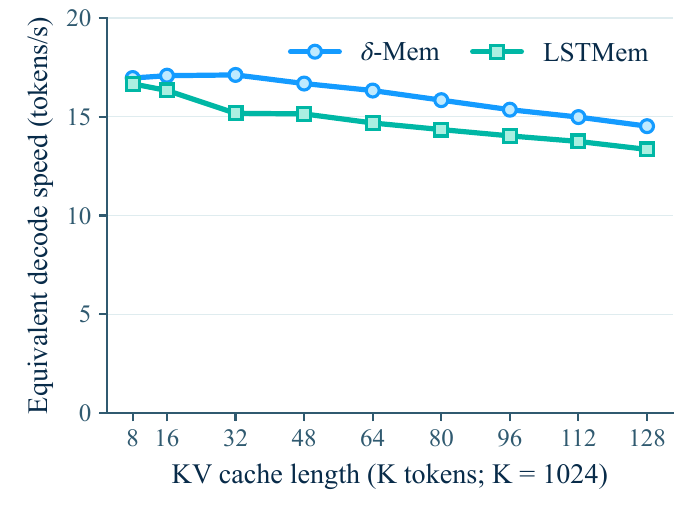}
\end{minipage}\hfill
\begin{minipage}[t]{0.49\linewidth}
\centering
\includegraphics[width=\linewidth]{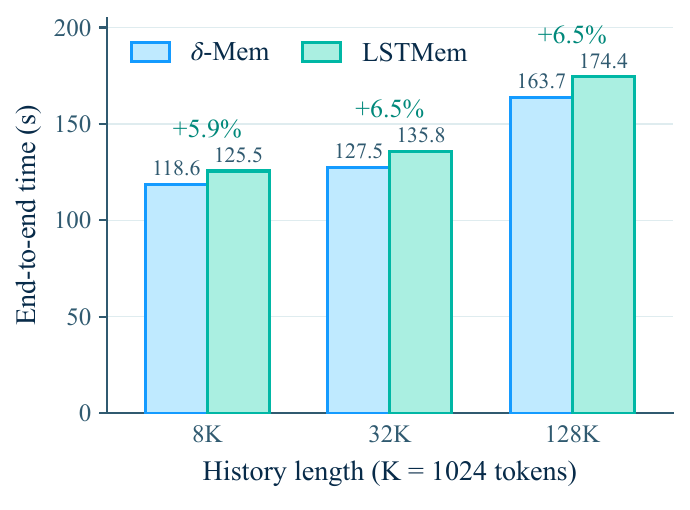}
\end{minipage}
\caption{Inference efficiency of \lstmem{} and \dmem{} on Qwen3-4B-Instruct with batch size one on H100 80GB GPUs. (Left) Equivalent decode speed with KV caches from 8K to 128K tokens, given by the reciprocal of the mean per-GPU median single-token latency. (Right) Mean end-to-end time for processing an 8K, 32K, or 128K-token history and generating a 2,048-token answer. Labels give the relative increase of \lstmem{} over \dmem{}, computed from unrounded times.}
\label{fig:efficiency}
\end{figure}
\section{Inference efficiency}
\label{app:efficiency}

\lstmem{} preserves most of \dmem{}'s inference efficiency, both for a single decoding step and for the complete pipeline from history processing to answer generation, with a modest incremental cost for its richer memory interface.
We compare the two methods on Qwen3-4B-Instruct, using the main configuration of \lstmem{}.

\textbf{Decoding speed.}
Both methods use batch size one, BF16, FlashAttention \citep{dao2024flashattention2}, and the same optimized inference runtime.
We evaluate nine KV-cache lengths from 8K to 128K tokens, where K denotes 1,024 tokens.
Each method runs independently on each of four NVIDIA H100 80GB GPUs, with model order alternated across devices.
For every model, cache length, and GPU, we collect 20 timed repetitions after two warm-up repetitions, yielding 1,440 measurements in total.
Each reported latency is the mean of the four per-GPU medians.
We time one synchronized token-decoding forward pass from restored KV and memory states.
For latency $t$ in milliseconds, the plotted \emph{equivalent decode speed} is $1000/t$ tokens/s.
This metric isolates the decoding step, excluding model loading, prefill, KV copying, memory-state restoration, warm-up, compilation, and token selection.
The resulting values characterize single-step decoding from an existing memory state.

\textbf{End-to-end cost.}
Because decoding speed excludes history processing, we also time the complete inference pipeline.
Each run starts from an empty memory state, writes a history of 8K, 32K, or 128K tokens in 2,048-token blocks, including the block-end feedback of \lstmem{}, and then prefills the question and generates a 2,048-token answer.
As in our evaluation protocol, the history KV cache remains available when answering, and model loading and warm-up are excluded.
Each configuration is timed 40 times per method on H100 80GB GPUs with batch size one, and we report the mean over all timings.

\textbf{Results.}
As shown in Figure~\ref{fig:efficiency} (left), \lstmem{} retains $92.10\%$ of \dmem{}'s equivalent decode speed on average, computed as the mean of the speed ratios across the nine cache lengths.
The corresponding mean relative latency increase is $8.68\%$.
At 128K tokens, \lstmem{} achieves $13.34$ tokens/s versus $14.52$ for \dmem{}, retaining $91.90\%$ of its speed with only $6.07$ ms of additional latency per step.
The added latency remains below $7.5$ ms at every evaluated length.
With 2,048-token answers (Figure~\ref{fig:efficiency}, right), \lstmem{} takes $125.5$, $135.8$, and $174.4$ seconds for 8K, 32K, and 128K histories, compared with $118.6$, $127.5$, and $163.7$ seconds for \dmem{}, so its end-to-end overhead is only $5.9\%$ to $6.5\%$.
Block-end feedback runs only while the history is processed, so its cost is amortized as the answer grows.
These timings include diagnostic statistics computed during feedback that are not required for inference.

\end{document}